\documentclass[runningheads]{llncs}
\usepackage[T1]{fontenc}
\usepackage{booktabs}
\usepackage{tabularx}
\usepackage[inline]{enumitem}
\usepackage{amsmath}
\usepackage{amsfonts}
\usepackage{xcolor}
\usepackage{graphicx}
\usepackage{subcaption}
\usepackage{multirow}
 \usepackage{hyperref}
\usepackage{graphicx}

\begin{document}
\title{GPT-FT: An Efficient Automated Feature Transformation Using GPT for Sequence Reconstruction and Performance Enhancement}

\author{Yang Gao\inst{1} \and
Dongjie Wang\inst{2} \and
Scott Piersall\inst{1} \and
Ye Zhang\inst{3} \and
Liqiang Wang\inst{1}}

\institute{University of Central Florida, Orlando FL 32816, USA  \\
\email{\{yang.gao,sc382961,liqiang.wang\}@ucf.edu}
\and University of Kansas, Lawrence KS 66045, USA \\
\email{wangdongjie@ku.edu}\\
\and Northeast Normal University, Changchun Jilin PRC \\ \email{zhangy923@nenu.edu.cn}\\}

\maketitle 
\pagestyle{empty}

\begin{abstract}
Feature transformation plays a critical role in enhancing machine learning model performance by optimizing data representations.
Recent state-of-the-art approaches address this task as a continuous embedding optimization problem, converting discrete search into a learnable process. 
Although effective, these methods often rely on sequential encoder-decoder structures that cause high computational costs and parameter requirements, limiting scalability and efficiency.
To address these limitations, we propose a novel framework that accomplishes automated feature transformation through four steps: transformation records collection, embedding space construction with a revised Generative Pre-trained Transformer (GPT) model, gradient-ascent search, and autoregressive reconstruction. In our approach, the revised GPT model serves two primary functions: (a). feature transformation sequence reconstruction; and (b) model performance estimation and enhancement for downstream tasks by constructing the embedding space. Such a multi-objective optimization framework reduces parameter size and accelerates transformation processes. 
Experimental results on benchmark datasets show that the proposed framework matches or exceeds baseline performance, with significant gains in computational efficiency.
This work highlights the potential of transformer-based architectures for scalable, high-performance automated feature transformation.
\end{abstract}

\section{Introduction}

Feature transformation is a pivotal component in machine learning pipelines, aiming to enhance downstream tasks' model performance by optimizing data representations. 
An effective feature transformation can significantly impact the predictive accuracy of models, especially in scenarios involving complex or high-dimensional datasets. 
Traditional approaches often rely on manual feature engineering, which is time-consuming and requires substantial domain expertise. 
This has spurred interests and studies in automated feature transformation (AFT) methods that can systematically and efficiently explore feature space.

The current algorithms for AFT can be broadly classified into three distinct categories:
(1) \textbf{Expansion-Reduction Methodologies}: These approaches, such as Deep Feature Synthesis~\cite{kanter2015deep}, AutoFeat~\cite{horn2020autofeat}, and Cognito~\cite{khurana2016cognito}, apply various mathematical operations across all features to generate a large set of potential transformed features, followed by a reduction phase to select the most valuable ones. 
Although these methods capture complex feature interactions, they often rely on random generation, leading to computational inefficiency and instability because of the inclusion of many redundant features.
(2) \textbf{Iterative-Feedback Approaches}: Methods like Group Feature Generation~\cite{wang2022group}, Feature Engineering Automation~\cite{olson2016evaluation,wu2025iterative}, and Genetic Programming-based techniques~\cite{tran2019genetic} combine feature generation and selection in iterative cycles, updating strategies based on model performance feedback. While guided by evolutionary algorithms or reinforcement learning, their reliance on discrete search spaces can hinder convergence, making scaling to larger feature spaces challenging and inefficient.
(3) \textbf{Neural Architecture Search (NAS)-Based Approaches}: Inspired by NAS, which was originally designed to automate neural network architecture design, some studies have framed AFT as a problem within the NAS paradigm~\cite{chen2019neural,zhu2022difer,wang2023reinforcement,liu2025continuous}. 
These methods treat feature transformation sequences as hyperparameters within a model structure for optimization.
Although this structured formulation guides the search, it often suffers from slow speeds and large parameter sizes, limiting efficiency and scalability.

To address these limitations, we introduce a novel, \textbf{G}enerative \textbf{P}re-trained \textbf{T}ransformer framework for efficient Automated \textbf{F}eature \textbf{T}ransformation (\textbf{GPT-FT}). 
Transformers provide strong sequence modeling and generation capabilities, enabling parallelization and improved parameter efficiency over traditional methods.
Our framework includes four key steps:
(1) \textbf{Transformation Record Collection}: We gather a dataset comprising feature transformation sequences and their corresponding model performance metrics. This dataset serves as the foundation for learning the relationship between transformation sequences and their impact on model performance.
(2) \textbf{Embedding Space Construction with revised GPT}: We adopt the architecture of the GPT-1 model~\cite{radford2018gpt} and train it from scratch to regenerate transformation sequences. Notably, our model, GPT-FT, is significantly smaller than GPT-1 in terms of parameter size, with an embedding size of 64 compared to GPT-1's 768. This step aims to two purposes: (a) \textit{feature transformation sequence reconstruction}, which learns to generate valid and effective transformation sequences in an autoregressive manner; (b) \textit{model performance estimation and optimization}, which predicts the performance impact of given transformation sequences to guide the optimization process.
(3) \textbf{Gradient-Ascent Search}: We perform optimization in the continuous embedding space constructed by our GPT-FT model. By applying gradient-ascent techniques, we efficiently search for embeddings that are likely to yield improved model performance.
(4) \textbf{Autoregressive Reconstruction}: The optimized embeddings are decoded back into feature transformation sequences using GPT-FT's autoregressive capabilities. This results in refined feature spaces tailored for enhanced downstream model performance.

By integrating sequence reconstruction and performance estimation/enhancement tasks within our decoder-only GPT-FT model, our approach significantly reduces parameter size and computational overhead compared to traditional encoder-decoder methods. This streamlined, decoder-only structure minimizes the parameter requirements, enhancing scalability and making the framework suitable for large-scale and real-time applications. 
We evaluate our framework on benchmark datasets, where it matches or surpasses state-of-the-art methods and achieves significant computational efficiency, highlighting the advantages of transformer-based architectures for automated feature transformation.

Our contributions can be summarized as follows:
\begin{itemize}
    \item We introduce a novel framework, GPT-FT, that leverages the GPT model architecture for efficient automated feature transformation, addressing the scalability and efficiency challenges present in existing methods.
    \item We show the dual capability of the GPT-FT model in reconstructing transformation sequences and estimating model performance within a unified architecture, enabling effective optimization in a continuous embedding space.
    \item We show through extensive experiments that our framework achieves superior performance with reduced computational costs compared to state-of-the-art methods.
\end{itemize}

\section{Problem Statement}

Our objective is to provide a resilient, deeply differentiable system for automatic feature transformation. Considering a dataset \( D = \{X, y\} \) and an operation set \( \mathcal{O} \), we develop a cascading reinforcement learning framework \( \tau \) to collect training data \( T = \{(\gamma_i, v_i)\}_{i=1}^n \), where \( \gamma_i \) denotes a sequence of feature transformations and \( v_i \) indicates its predictive performance. Our framework concurrently optimises a mapping function \( \phi \), a reconstruction function \( \psi \), and an evaluation function \( \omega \) to embed transformation sequences into a continuous space, linking each point with its corresponding sequence and performance metrics. Through gradient-based search in the embedding space, we determine the best transformation sequence \( \Gamma^* \), which may be expressed as follows:

\begin{equation}
\Gamma^* = \psi(\textbf{E}^*) = \arg\max_{\textbf{E}}\mathcal{P}(\mathcal{Q}(\psi\{\phi[\tau(X)]\}), y), 
\end{equation}
where \( \tau \) transforms the original dataset feature \( X \) into \( \{\gamma_i\}^n_{i=1} \), \( \phi \) maps \( \{\gamma_i\}^n_{i=1} \) to a continuous embedding space, and \( \psi \) reconstructs a sequence of feature transformations from any embedding point; \( \textbf{E}^* \) denotes the optimal embedding; \( \mathcal{Q} \) represents the downstream machine learning model; and \( \mathcal{P} \) indicates the performance metric. Ultimately, we employ \( \Gamma^* \) to convert \( \textbf{X} \) into the optimal feature space \( \textbf{X}^* \), therefore maximising \( \mathcal{P} \).

\section{Methodology}
\subsection{Framework Overview}

\begin{figure}[!t]
    \centering
    \includegraphics[width=1\textwidth]{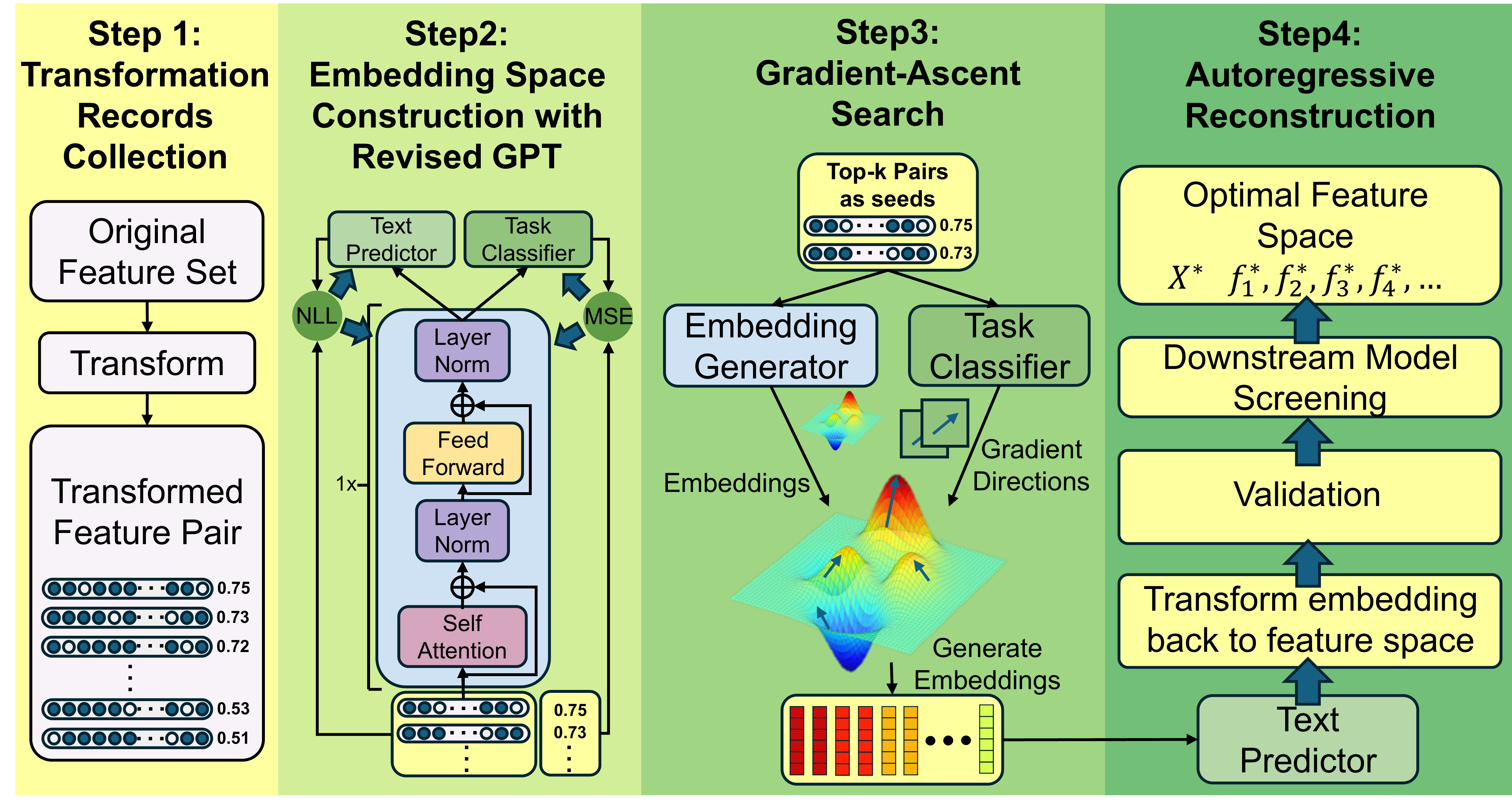} 
    \caption{An overview of our model framework GPT-FT.}
    \label{fig:model}
    \vspace{-0.5cm}
\end{figure}


Figure~\ref{fig:model} shows the framework of GPT-FT with four steps: 
\begin{enumerate*}[label=(\arabic*)]
    \item Transformation Records Collection.
    \item Embedding Space Construction with a revised GPT.
    \item Gradient-Ascent Search.
    \item Autoregressive Reconstruction.
\end{enumerate*}
\textbf{In Step 1}, we collect records of feature transformation sequences and their associated model performance using an RL-based framework, as described in ~\cite{wang2022group}.
\textbf{In Step 2}, our GPT-FT model  encodes knowledge from the collected feature transformation records into a continuous embedding space.
To achieve this, we minimize both the feature transformation sequence reconstruction loss and the model performance estimation loss.
\textbf{In Step 3}, we initially acquire the embeddings of the highest-ranking transformation operation sequences through the well-trained GPT-FT. Using these embeddings as initial points, we explore the gradient generated by the GPT-FT to identify optimal embeddings that enhance model performance. 
\textbf{In Step 4}, the GPT-FT based Text Predictor decodes optimal embeddings to generate candidate feature transformation sequences. These sequences are applied to the original features to construct refined feature spaces. A downstream predictive model evaluates the quality of these spaces, and the feature space with the highest performance is selected as the optimal output.

\subsection{Transformation Record Collection}
To automatically collect a large volume of high-quality transformation records, we employ an RL-based feature transformation framework as data collector~\cite{wang2022group}.
Specifically, the feature transformation process is modeled as three Markov Decision Processes (MDPs): a head feature agent, an operation agent, and a tail feature agent. These agents work collaboratively to select candidate features and mathematical operations for generating new features. The process is optimized to maximize downstream predictive performance while minimizing feature space redundancy. During this learning phase, transformation sequences and their corresponding model performance are collected to prepare data $T=\{(\gamma_i, v_i)\}^n_{i=1}$ where $\gamma_i$ is transformed feature sequence, $v_i$ is the corresponding downstream task performance and $n$ is the number of the pairs.
\label{sec:transrec}

\subsection{Embedding Space Construction with GPT}
\label{sec:gpt}
We use GPT-FT to map the sequential information of preprocessed features into an embedding space. Each feature is represented as a pair of a transformation operation sequence and its corresponding model performance. GPT-FT produces two outputs—text-based predictions (\( \hat{\gamma}_i \)) and downstream task performance (\( \hat{v}_i \))—leading to two distinct training objectives.

\noindent\textbf{Target 1: Learning Continuous Embeddings.}  
The first objective is to train GPT-FT to generate continuous embeddings that effectively represent the original dataset while reducing the search space. These continuous embeddings can be explored using gradient-based optimization. To achieve this, our GPT-FT uses the single-layer \textit{Embedding Generator \( \phi \)} (reduced from the original 12 layers in GPT ~\cite{radford2018gpt}). We train the \textit{Embedding Generator \( \phi \)} alongside the \textit{Text Predictor \( \psi \)},
both utilizing the same input-output pairs transformed in Step 1. The forward process is expressed as \( \hat{\gamma}_i =  \psi (\textbf{E}_i) \), 
where \( \textbf{E}_i = \phi(\gamma_i) \in \mathbb{R}^{L \times d} \), 
and \( L \) denotes the length of \( \gamma_i \) and \( \hat{\gamma}_i \).
For the loss function, we assume the GPT-FT output follows a probability distribution centered on the input sequence \( \gamma_i \) with unit variance. Accordingly, we employ the Negative Log-Likelihood (NLL) loss: \( \mathcal{L}_{\text{pre}} = \sum_{i=1}^{n} -\log p(\hat{\gamma}_i | \gamma_i) \), where \( \hat{\gamma}_i \) is the GPT-FT's text-based output, \( \gamma_i \) is the input sequence, and \( n \) is the number of feature-performance pairs as defined in Section~\ref{sec:transrec}.

\noindent\textbf{Target 2: Estimating Downstream Task Performance.}  
The second objective is to train GPT-FT to estimate the downstream task performance, enabling gradient-based guidance for subsequent search steps. Here, the ground truth is the model performance \( v_i \) (e.g., F1-score or \( 1 - \text{RAE} \)) from Step 1. We train the single-layer \textit{Embedding Generator \( \phi \)} alongside the \textit{Task Classifier \( \delta \)} in GPT-FT to predict performance values, formulated as \( \hat{v}_i = \delta(\textbf{E}_i) \in \mathbb{R} \).
The loss function for this objective is defined as \( \mathcal{L}_{\text{cls}} = \sum_{i=1}^{n} \text{MSE}(\hat{v}_i, v_i) \), where \( \hat{v}_i \) is \textit{Task Classifier \( \delta \)}'s predicted performance, and \( v_i \) is the actual performance from Step 1.
Both the \textit{Text Predictor} and \textit{Task Classifier} are implemented as single-layer linear transformations.

\noindent\textbf{Joint Training Loss $\mathcal{L}$:}
We jointly optimize the GPT-FT model. The joint training loss can be formulated as: $\mathcal{L} = \alpha \mathcal{L}_{pre} + (1-\alpha)\mathcal{L}_{cls}$, where $\alpha$ is the trade-off hyperparameter that controls the contribution of sequence reconstruction and accuracy estimation loss.
\label{sec:gpt:joint}

\subsection{Gradient-Ascent Search}
To perform optimal embedding search, we first select the top-\( k \) transformation sequences ranked by downstream predictive accuracy. The trained GPT-FT maps these postfix expressions to continuous embeddings, which serve as initial points for gradient ascent. Starting from an embedding \( \textbf{E} \), the search updates as \( \tilde{\textit{E}} = \textbf{E} + \eta \frac{\partial \textbf{G}}{\partial \textbf{E}} \), where \( \tilde{\textbf{E}} \) is the refined embedding, \( \eta \) is the step size, and \( \textbf{G} \) represents GPT-FT. The performance satisfies \( \textbf{G}(\tilde{\textbf{E}}) \geq \textbf{G}(\textbf{E}) \). For \( k \) seeds, the refined embeddings are \( [\tilde{\textbf{E}}_1, \tilde{\textbf{E}}_2, \dots, \tilde{\textbf{E}}_k] \).

\subsection{Autoregressive Reconstruction}
The trained \textit{Text Predictor \( \psi \)} in GPT-FT reconstructs transformation sequences from the candidate embeddings \( [\tilde{\textbf{E}}_1, \tilde{\textbf{E}}_2, \dots, \tilde{\textbf{E}}_k] \) as \( [\tilde{\textbf{E}}_1, \dots, \tilde{\textbf{E}}_k] \xrightarrow{\psi} \{\tilde{\gamma}_i\}_{i=1}^k \). The sequence with the highest probability is selected, generating \( k \) transformation sequences \( \{\tilde{\gamma}_i\}_{i=1}^k \). Each sequence is segmented by the <SEP> token, with invalid segments removed based on mathematical computability. Valid components reconstruct feature transformation sequences \( \{\tilde{\Gamma}_i\}_{i=1}^k \), which refine the feature space \( \{\tilde{X}_i\}_{i=1}^k \). The feature set yielding the highest downstream performance is selected as the optimal feature space \( \textbf{X}^* \).

\section{Experiment}

\subsection{Experimental Setup}

\noindent\textbf{Datasets and Evaluation Metrics}
We conducted experiments on 15 publicly available datasets from Kaggle~\cite{howard_kaggle_2022}, OpenML~\cite{openml_2022}, and UCI~\cite{uci_ml_repo}, comprising nine classification and six regression tasks. Dataset statistics are summarized in Table~\ref{tab:expperf}. For classification tasks, we used F1-score, Precision, Recall, and ROC/AUC, while regression tasks were evaluated using 1-Relative Absolute Error (1-RAE)~\cite{wang2022group}, 1-Mean Absolute Error (1-MAE), 1-Mean Square Error (1-MSE), and 1-Root Mean Square Error (1-RMSE). 

\noindent\textbf{Baseline Models}
We compared our method against nine prevalent feature generation techniques:  
(1) RDG generates transformation records of feature-operation-feature randomly to create a new feature space.  
(2) ERG applies operations to each feature to expand the feature space, then selects essential features as the new feature set.  
(3) LDA~\cite{blei2003lda} employs matrix factorization to derive hidden states as the generated feature space.  
(4) AFAT~\cite{horn2020autofeat} improves upon ERG by iteratively generating new features and using multi-step feature selection to identify informative ones.  
(5) NFS~\cite{chen2019neural} models transformation sequences for each feature and optimizes feature generation using reinforcement learning.  
(6) TTG~\cite{khurana2018feature} conceptualizes the transformation process as a graph and applies reinforcement learning to search for the optimal feature set.  
(7) GRFG~\cite{wang2022group} uses three collaborative reinforced agents for feature generation and introduces feature grouping to improve learning efficiency.  
(8) DIFER~\cite{zhu2022difer} employs a seq2seq model to embed randomly generated feature transformations and applies gradient search to identify optimal features.  
(9) MOAT~\cite{wang2023reinforcement} uses an embedding-optimization-reconstruction framework to reformulate discrete feature transformations as a continuous optimization task, leveraging an encoder-evaluator-decoder structure to enhance data utilization from GRFG.

\begin{table}[htbp]
  \centering
  \rotatebox{90}{ 
    \begin{minipage}{\textheight} 
      \centering
      \caption{Comparison of Overall Performance: Results for binary classification are labeled as "C," while "R" indicates regression tasks. The highest performance values are shown in \textbf{bold}, with the second-highest values underlined. (\textbf{Greater values signify superior performance.})}
      {\small
        \begin{tabularx}{\textwidth}{cccccccccccccccc}
        \toprule
        \textbf{Dataset} & \textbf{Source} & \textbf{C/R} & \textbf{Samples} & \textbf{Features} & \textbf{RDG} & \textbf{ERG} & \textbf{LDA} & \textbf{AFAT} & \textbf{NFS} & \textbf{TTG} & \textbf{GRFG} & \textbf{DIFER} & \textbf{MOAT} & \textbf{GPT-FT} \\
        \midrule
        \parbox{2.5cm}{\centering Contraceptive \\ Method Choice~\cite{contraceptive_method_choice_30}} & UCIrvine & C & 1473 & 9 & 0.493 & 0.505 & 0.366 & 0.503 & 0.52 & 0.508 & 0.533 & \underline{0.538} & 0.537 & \textbf{0.544} \\
        \midrule
        Heart Disease~\cite{heart_disease_45} & UCIrvine & C & 303 & 13 & 0.851 & 0.850 & 0.763 & 0.834 & 0.834 & \textbf{0.868} & 0.831 & 0.841 & 0.866 & \underline{0.867} \\
        \midrule
        \parbox{2.5cm}{\centering Ozone Level \\ Detection~\cite{ozone_level_detection_172}}  & UCIrvine & C & 2536 & 72 & 0.959 & 0.959 & 0.957 & 0.961 & 0.956 & 0.96 & 0.958 & 0.956 & \underline{0.961} & \textbf{0.962} \\
        \midrule
        Seeds~\cite{seeds_236} & UCIrvine & C & 210 & 7 & 0.969 & 0.971 & 0.736 & 0.971 & 0.971 & 0.965 & 0.926 & 0.957 & \underline{0.971} & \textbf{0.977} \\
        \midrule
        Titanic~\cite{kaggle_titanic} & Kaggle & C & 891 & 11 & 0.814 & 0.829 & 0.736 & 0.818 & 0.82 & 0.814 & 0.82 & 0.825 & \underline{0.831} & \textbf{0.832} \\
        \midrule
        Lymphography~\cite{lymphography_63} & UCIrvine & C & 148 & 18 & 0.108 & 0.144 & 0.167 & 0.15 & 0.152 & 0.148 & 0.182 & 0.15 & \underline{0.267} & \textbf{0.352} \\
        \midrule
        Amazon Employee~\cite{amazon_employee_dataset} & Kaggle & C & 32769 & 9 & 0.932 & 0.934 & 0.916 & 0.93 & 0.932 & 0.933 & 0.932 & 0.929 & \underline{0.936} & \textbf{0.983} \\
        \midrule
        Wine Quality Red~\cite{wine_quality_186} & UCIrvine & C & 999 & 12 & 0.466 & 0.461 & 0.433 & 0.48 & 0.462 & 0.467 & 0.47 & 0.476 & \underline{0.559} & \textbf{0.622} \\
        \midrule
        Wine Quality White~\cite{wine_quality_186} & UCIrvine & C & 4900 & 12 & 0.524 & 0.510 & 0.449 & 0.516 & 0.525 & 0.507 & 0.534 & 0.507 & \underline{0.536} & \textbf{0.544} \\
        \midrule
        Tecator~\cite{tecator_dataset} & OpenML & R & 240 & 125 & 0.541 & 0.584 & 0.418 & 0.541 & 0.525 & 0.527 & \underline{0.750} & 0.692 & 0.545 & \textbf{0.885} \\
        \midrule
    \parbox{2.5cm}{\centering Geographical \\ OriginalofMusic~\cite{geographical_origin_of_music_315}} & UCIrvine & R & 1059 & 118 & 0.388 & 0.395 & 0.317 & 0.398 & 0.283 & 0.28 & 0.472 & \underline{0.632} & 0.481 & \textbf{0.508} \\
        \midrule
        Jasmine~\cite{jasmine_dataset} & OpenML & R & 2984 & 145 & 0.402 & 0.415 & 0.391 & 0.411 & 0.406 & 0.407 & 0.326 & \underline{0.447} & 0.407 & \textbf{0.477} \\
        \midrule
        Libras move~\cite{libras_movement_181} & OpenML & R & 360 & 91 & 0.179 & 0.286 & 0.085 & 0.215 & 0.156 & 0.226 & \underline{0.294} & 0.172 & 0.293 & \textbf{0.308} \\
        \midrule
        Bodyfat~\cite{johnson_bodyfat} & UCIrvine & R & 252 & 15 & 0.84 & 0.84 & 0.282 & 0.846 & 0.848 & \underline{0.853} & 0.652 & 0.737 & 0.843 & \textbf{0.865} \\
        \midrule
        Weather~\cite{weather_dataset} & Kaggle & R & 366 & 12 & 0.969 & 0.971 & 0.838 & 0.975 & 0.975 & 0.973 & 0.96 & 0.914 & \underline{0.976} & \textbf{0.98} \\
        \bottomrule
        \end{tabularx}
      }
      \label{tab:expperf}
    \end{minipage}
  }
\end{table}

\noindent\textbf{Experimental Platform}
To evaluate GPT-FT against baseline models, we present the results of quantitative and qualitative experiments. All experiments were conducted on an Intel Xeon Silver 4114 CPU and four NVIDIA TITAN RTX GPUs. Additional platform details are provided in Appendix~\ref{apx:pltfm}.

\begin{figure}[!htbp]
    \centering
    \includegraphics[width=1\textwidth]{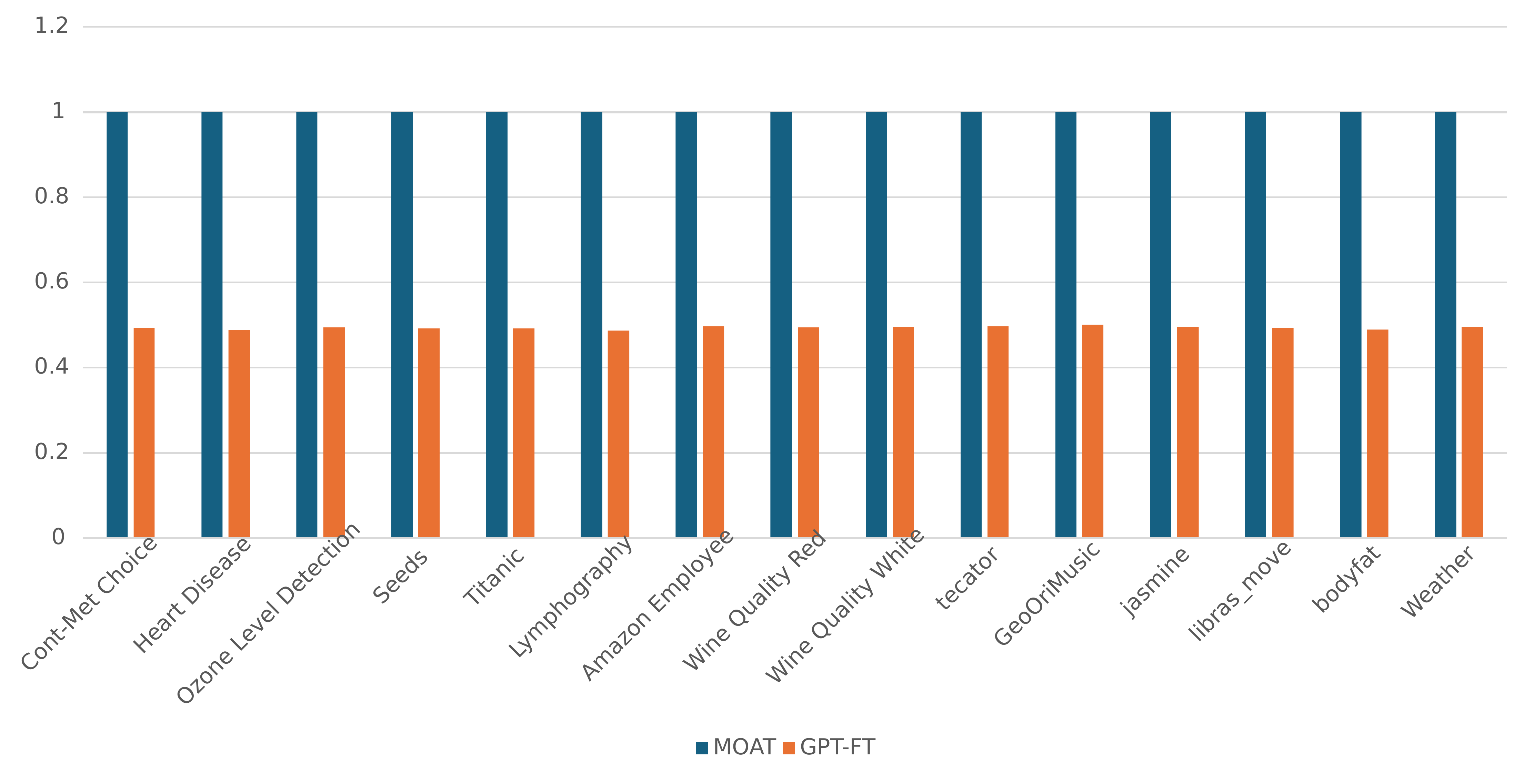} 
    \vspace{-0.5cm}
    \caption{Comparison of parameter size across different datasets. We normalize the bigger value to 1 and less value to [0,1].}
    \label{fig:modelsize}
    \vspace{-1.6cm}
\end{figure}

\begin{figure}[!htbp]
    \centering
    \includegraphics[width=1\textwidth]{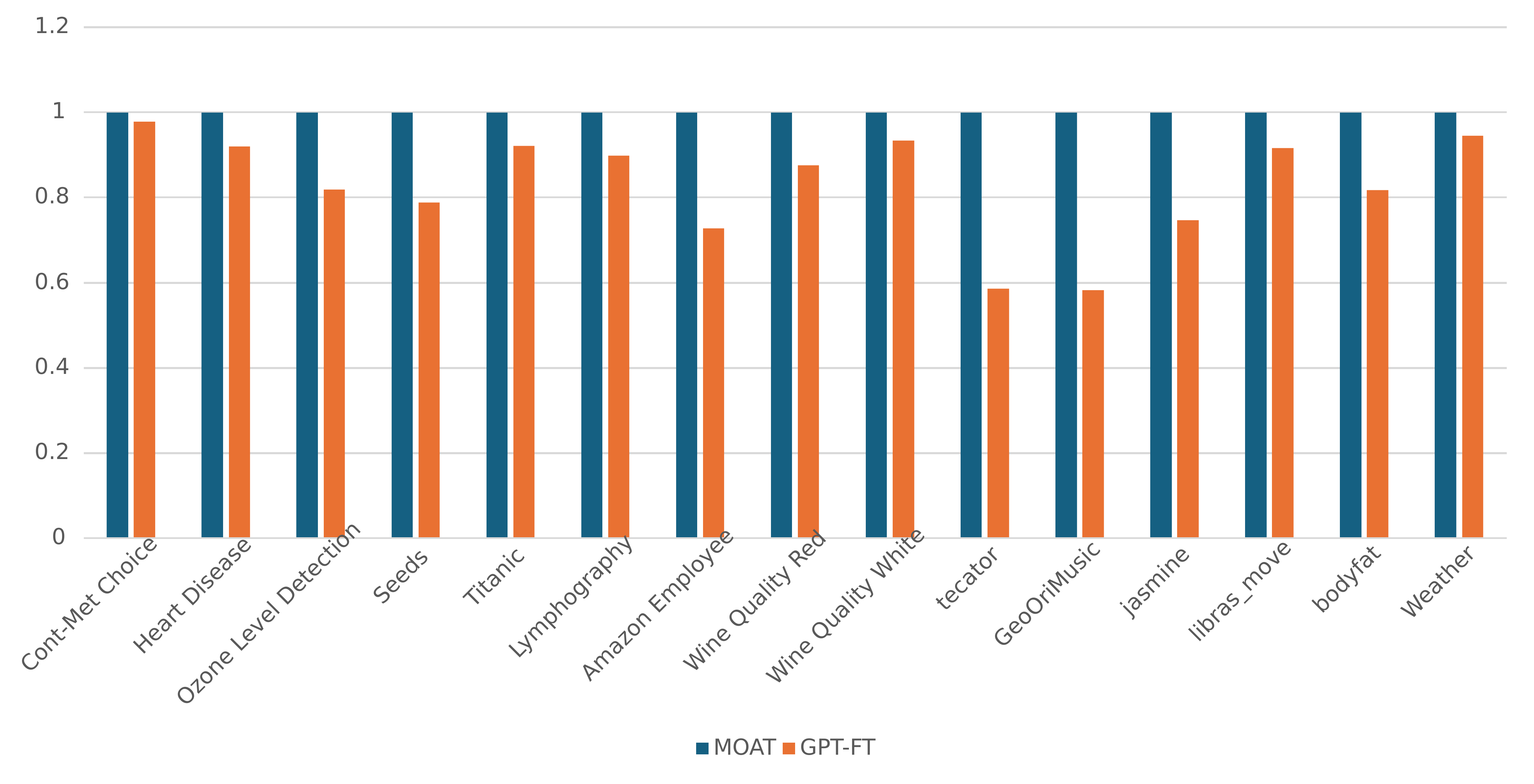} 
    \vspace{-0.4cm}
    \caption{Comparison of inference times across different datasets. We normalize the bigger value to 1 and less value to [0,1].}
    \label{fig:infertime}
    \vspace{-1.2cm}
\end{figure}

\subsection{Performance Evaluation}
\noindent\textbf{Overall Performance.}
This experiment evaluates \textit{GPT-FT's ability to generate transformation sequences for identifying an optimal feature space with superior performance}. Table~\ref{tab:expperf} compares GPT-FT with other models on F1-score and 1-RAE metrics, showing GPT-FT outperforms all others across datasets. GPT-FT's efficient embedding space preserves feature transformation knowledge, enabling its gradient-ascent module to locate the optimal feature space effectively. Compared to MOAT, GPT-FT achieves better stability due to: 1) RL-based data collection providing a solid foundation for a discriminative embedding space; 2) Postfix notation reducing the search space, improving transformation knowledge acquisition. This reflects GPT-FT's efficacy.

\noindent\textbf{Inference Time and Parameter Size.}
To facilitate a clear comparison of inference time and parameter size, we normalize their values to the range [0,1] using the min-max normalization approach for each dataset, with comprehensive values included in Appendix~\ref{apx:expdtl}. 
Figure~\ref{fig:modelsize} shows GPT-FT consistently has smaller parameter sizes than MOAT across datasets, indicating greater design efficiency. For example, in the Amazon Employee dataset, GPT-FT’s size is 3.21 MB versus MOAT’s 6.46 MB (a~50\% reduction), and in the Geographical Origin of Music dataset, GPT-FT uses 7.08 MB compared to MOAT’s 14.14 MB. Even in smaller datasets like Heart Disease, GPT-FT (0.10 MB) remains more compact than MOAT (0.20 MB). Figure~\ref{fig:infertime} compares inference times, where GPT-FT consistently outperforms MOAT. For instance, in the Ozone Level Detection dataset, GPT-FT achieves an 18\% improvement (27.93s vs. 34.11s), and in the Tecator dataset, it reduces inference time by 41\% (39.42s vs. 67.22s). Even in smaller datasets like Heart Disease, GPT-FT (22.61s) is faster than MOAT (24.58s). These results highlight GPT-FT’s efficiency in both parameter size and inference speed, making it a strong choice for applications requiring optimized performance.

\noindent\textbf{Robustness Check.}
This experiment evaluates \textit{GPT-FT's robustness across various downstream machine learning models}. We tested Random Forest (RF), XGBoost (XGB), Support Vector Machine (SVM), K-Nearest Neighbors (KNN), Ridge, LASSO, and Decision Tree (DT), with results for Weather and Wine Quality Red datasets shown in Table~\ref{tab:exprobwea} and Table~\ref{tab:exprobwine}, using 1-RAE and F1-score metrics, respectively. GPT-FT consistently beats MOAT across models, likely due to its RL-based data collector tailoring transformation records to the downstream model. The embedding space effectively captures model-specific characteristics, enabling optimal feature space generation. These results highlight GPT-FT's robustness.

\begin{figure}[h]
    \centering
    \begin{subfigure}[b]{0.45\textwidth}
        \centering
        \includegraphics[width=\linewidth]{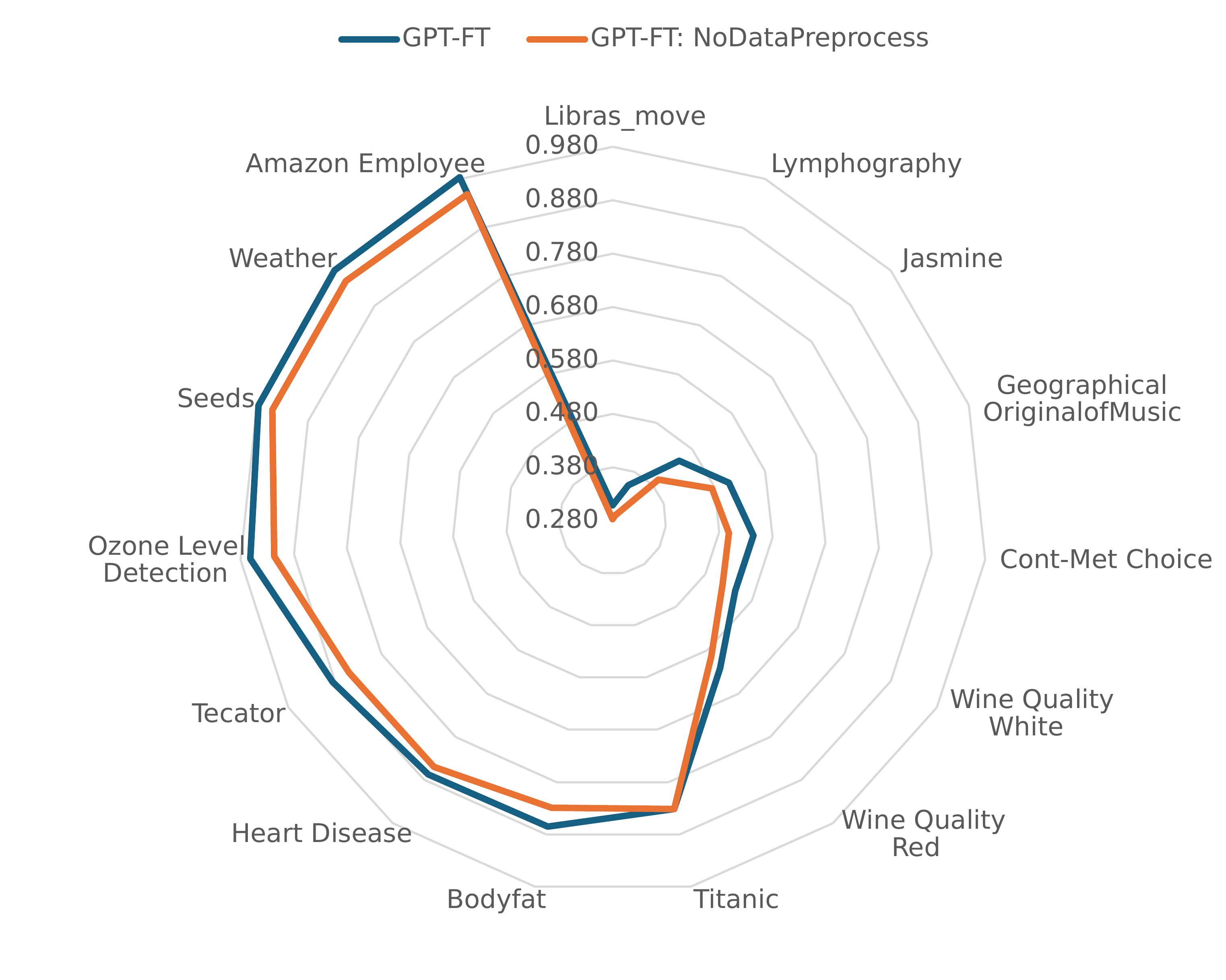}
        \caption{Comparison of performance across all datasets. Data pre-processing consistently outperforms the absence of such techniques across all data sets, albeit to varying extents.}
        \label{fig:nodata}
    \end{subfigure}
    \hfill
    \begin{subfigure}[b]{0.45\textwidth}
        \centering
        \includegraphics[width=\linewidth]{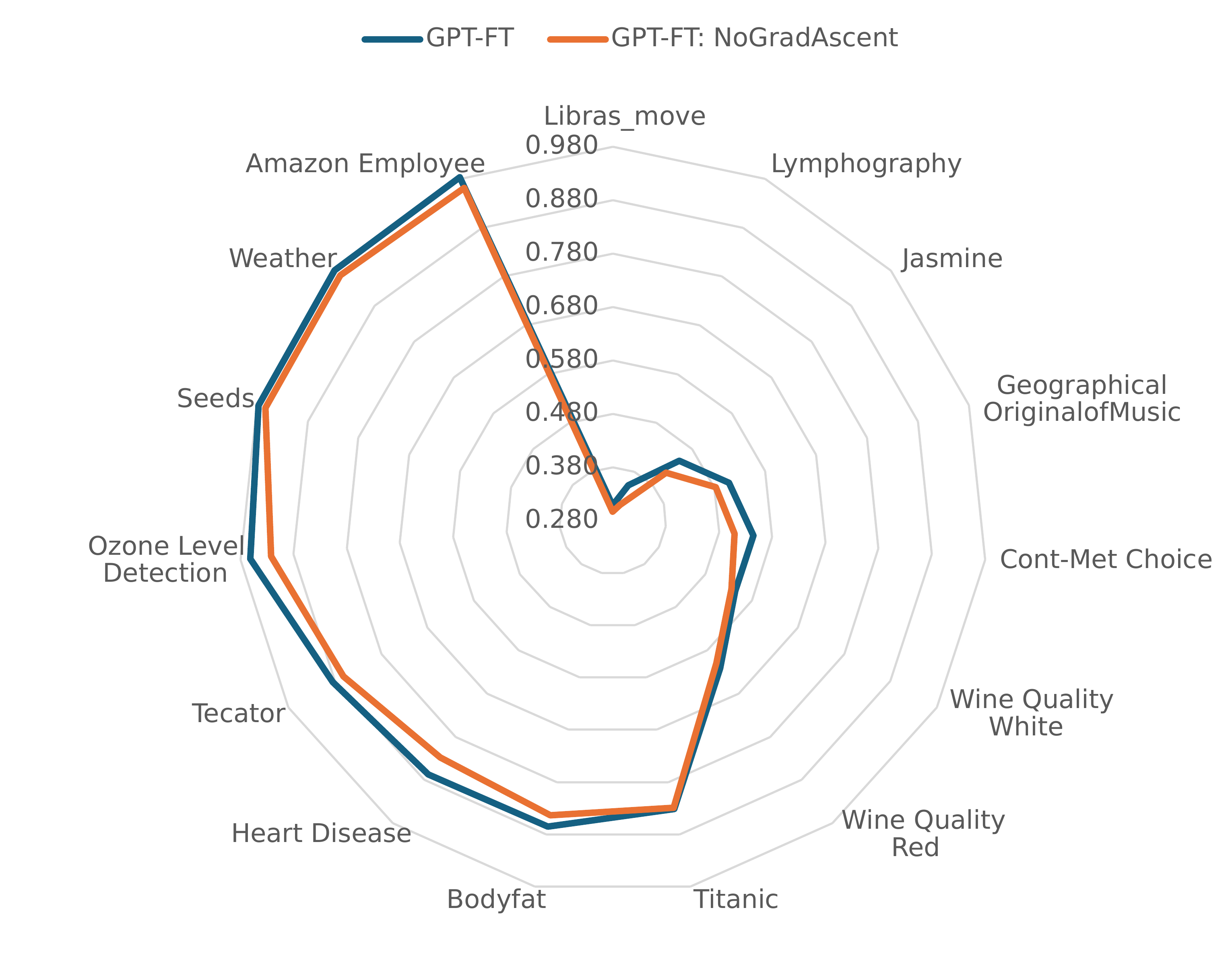}
        \caption{Comparison of performance across all datasets. Gradient ascent consistently outperforms the method without gradient ascent across all data sets, albeit to varying degrees.}
        \label{fig:nograd}
    \end{subfigure}
    \caption{Comparison of performance across datasets, considering (a) data pre-processing and (b) gradient ascent.}
    \label{fig:combined}
\end{figure}

\noindent\textbf{Ablation Study.}
To assess the impacts of Step 1: transformation records collection and Step 3: gradient ascent search in GPT-FT, we executed two experiments. Figure~\ref{fig:nodata} illustrates the outcomes devoid of Step 1 (gathering of transformation records), whereby the original dataset substitutes the altered feature collection. Step 1 enhances performance in the Contraceptive Method Choice and Weather datasets but has little effect on the Titanic dataset, possibly because of the simplicity of Titanic's characteristics, while the additional features in Step 1 facilitate GPT-FT's acquisition of more complicated information in the other datasets. 
Figure~\ref{fig:nograd} displays outcomes excluding Step 3 (gradient ascent search), with the gradient step established at 0. In the absence of Step 3, performance markedly declines in Contraceptive Method Choice and Weather, while seeing just a little reduction in Titanic. The embedding space for Titanic is probably near-optimal with small gradients, but greater gradients in the other datasets substantially enhance GPT-FT's performance.

\begin{table}[!ht]
    \centering
      \caption{Robustness check of GPT-FT with distinct ML models on Weather dataset in terms of 1-RAE score.}
    \begin{tabular}{cccccccc}
    \hline
        \textbf{Weather} & \textbf{RF} & \textbf{XGB} & \textbf{SVM} & \textbf{KNN} & \textbf{Ridge} & \textbf{LASSO} & \textbf{DT} \\ \hline
        RDG & 0.969 & 0.977 & 0.609 & 0.871 & 0.481 & 0.163 & 0.971 \\ \hline
        ERG & 0.971 & 0.971 & 0.722 & 0.862 & 0.48 & 0.104 & 0.973 \\ \hline
        LDA & 0.838 & 0.915 & 0.248 & 0.824 & 0.016 & 0.217 & 0.904 \\ \hline
        AFAT & 0.975 & 0.971 & 0.629 & 0.854 & 0.474 & 0.209 & 0.976 \\ \hline
        NFS & 0.975 & 0.974 & 0.614 & 0.865 & 0.202 & 0.132 & 0.976 \\ \hline
        TTG & 0.975 & 0.97 & 0.571 & 0.873 & 0.197 & 0.198 & 0.978 \\ \hline
        GRFG & 0.96 & 0.962 & 0.826 & 0.928 & 0.327 & 0.231 & 0.962 \\ \hline
        DIFER & 0.914 & 0.906 & 0.712 & 0.9 & 0.461 & 0.217 & 0.905 \\ \hline
        MOAT & 0.976 & 0.975 & 0.314 & 0.976 & 0.484 & 0.244 & 0.976 \\ \hline
        GPT-FT & \textbf{0.98} & \textbf{0.98} & \textbf{0.831} & \textbf{0.978} & \textbf{0.493} & \textbf{0.251} & \textbf{0.981} \\ \hline
    \end{tabular}
    \label{tab:exprobwea}
\end{table}

\begin{table}[!ht]
    \centering
    \caption{Robustness check of GPT-FT with distinct ML models on Wine Quality Red dataset in terms of F1-score.}
    \begin{tabular}{cccccccc}
    \hline
        \textbf{Wine Quality Red} & \textbf{RF} & \textbf{XGB} & \textbf{SVM} & \textbf{KNN} & \textbf{Ridge} & \textbf{LASSO} & \textbf{DT} \\ \hline
        RDG & 0.466 & 0.591 & 0.568 & 0.530 & 0.561 & 0.575 & 0.522 \\ \hline
        ERG & 0.461 & 0.574 & 0.570 & 0.561 & 0.557 & 0.576 & 0.515 \\ \hline
        LDA & 0.433 & 0.564 & 0.537 & 0.493 & 0.537 & 0.537 & 0.535 \\ \hline
        AFAT & 0.480 & 0.564 & 0.356 & 0.436 & 0.522 & 0.509 & 0.490 \\ \hline
        NFS & 0.462 & 0.561 & 0.559 & 0.530 & 0.573 & 0.583 & 0.468 \\ \hline
        TTG & 0.467 & 0.585 & 0.560 & 0.540 & 0.560 & 0.575 & 0.532 \\ \hline
        GRFG & 0.470 & 0.581 & 0.580 & 0.587 & 0.570 & 0.580 & 0.587 \\ \hline
        DIFER & 0.476 & 0.576 & 0.538 & 0.538 & 0.587 & 0.587 & 0.516 \\ \hline
        MOAT & 0.616 & 0.595 & 0.507 & 0.526 & 0.591 & 0.586 & 0.559 \\ \hline
        GPT-FT & \textbf{0.622} & \textbf{0.596} & \textbf{0.599} & \textbf{0.587} & \textbf{0.593} & \textbf{0.598} & \textbf{0.594} \\ \hline
    \end{tabular}
    \label{tab:exprobwine}
\end{table}

\begin{figure}[h]
    \centering
    \begin{subfigure}[b]{0.49\textwidth}
        \centering
        \includegraphics[width=\linewidth]{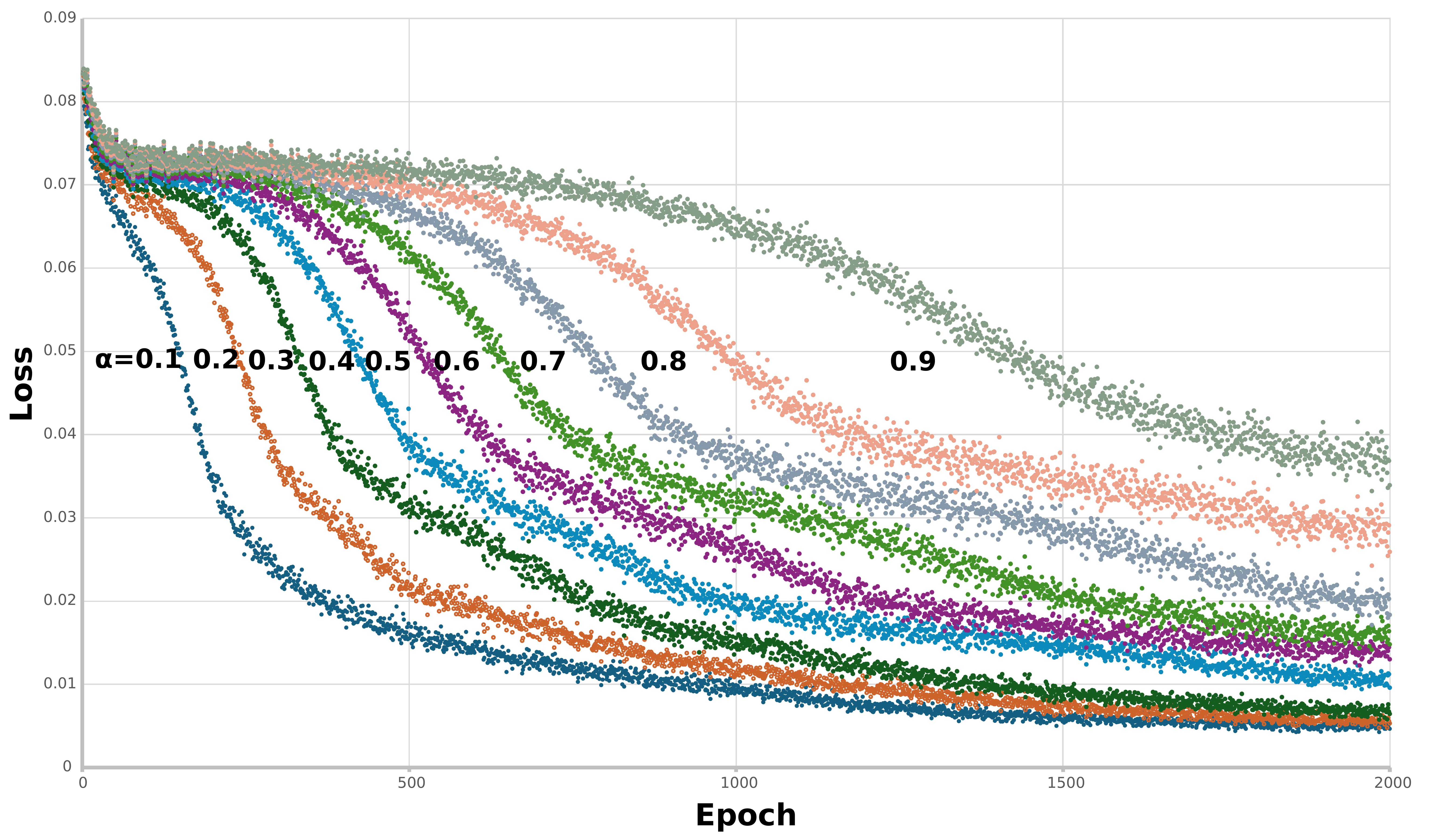}
        \caption{Trends in $\mathcal{L}_{cls}$ when training. From left to right is when $\alpha=0.1,0.2,...,0.9$.}
        \label{fig:clsloss}
    \end{subfigure}
    \hfill
    \begin{subfigure}[b]{0.49\textwidth}
        \centering
        \includegraphics[width=\linewidth]{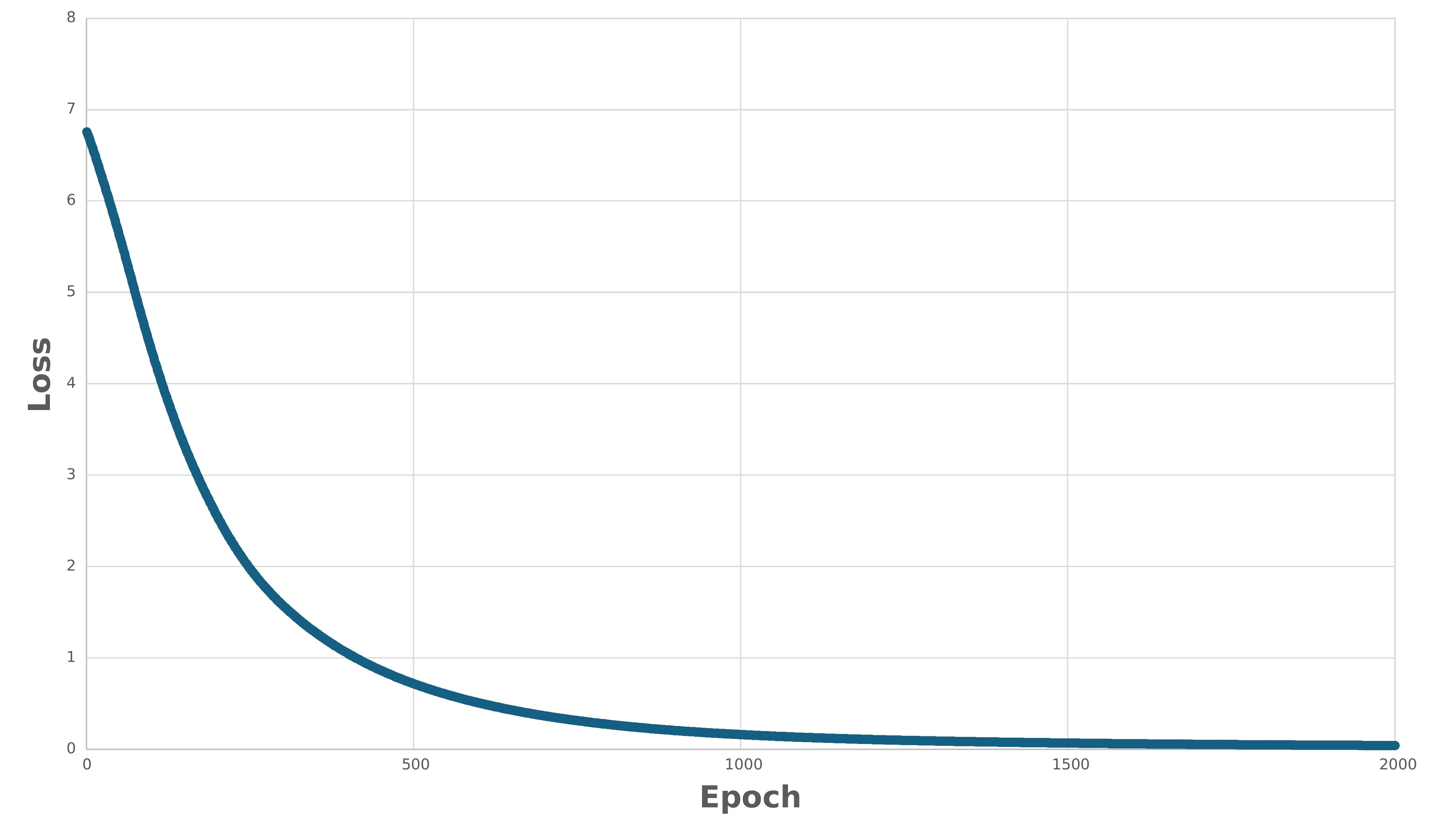}
        \caption{Trends in $\mathcal{L}_{pre}$ when training. Trends keep consistent when $\alpha=0.1,0.2,...,0.9$.}
        \label{fig:preloss}
    \end{subfigure}
    \caption{Trends in $\mathcal{L}_{cls}$ and $\mathcal{L}_{pre}$ during training for $\alpha = 0.1, 0.2, \ldots, 0.9$, demonstrating consistent behavior across different values of $\alpha$.}
    \label{fig:twoloss}
\end{figure}

\noindent\textbf{Parameter Sensitivity $\alpha$.}
To validate the sensitivity of the trade-off parameter \( \alpha \) in $\mathcal{L} = \alpha \mathcal{L}_{pre} + (1-\alpha)\mathcal{L}_{cls}$ (see Section~\ref{sec:gpt:joint}), we varied \( \alpha \) from 0.1 to 0.9 to observe its impact on training and performance. Lower \( \alpha \) reduces the contribution of sequence reconstruction loss \( \mathcal{L}_{pre} \) while allocating more gradient to accuracy estimation loss \( \mathcal{L}_{cls} \). Figure~\ref{fig:clsloss} shows \( \mathcal{L}_{cls} \) is highly sensitive to \( \alpha \); lower \( \alpha \) leads to faster convergence, while high \( \alpha \) (e.g., 0.9) causes a training barrier, delaying or preventing convergence. Meanwhile, it \( \mathcal{L}_{pre} \) consistently decreases regardless of \( \alpha \), reaching a low value after 1000 epochs (Figure~\ref{fig:preloss}). However, if training stops here, the target for \( \mathcal{L}_{cls} \) is unfilled, providing poor gradients for subsequent steps. Performance-wise, \( \alpha \in [0.4, 0.9] \) fails to generate valid records, so we restricted \( \alpha \) to [0.1, 0.3] and optimized it using NNI~\cite{nni2021}, setting \( \alpha = 0.133 \) as the best value.

\begin{figure}[h]
    \centering
    \begin{subfigure}[b]{0.49\textwidth}
        \centering
        \includegraphics[width=0.85\linewidth]{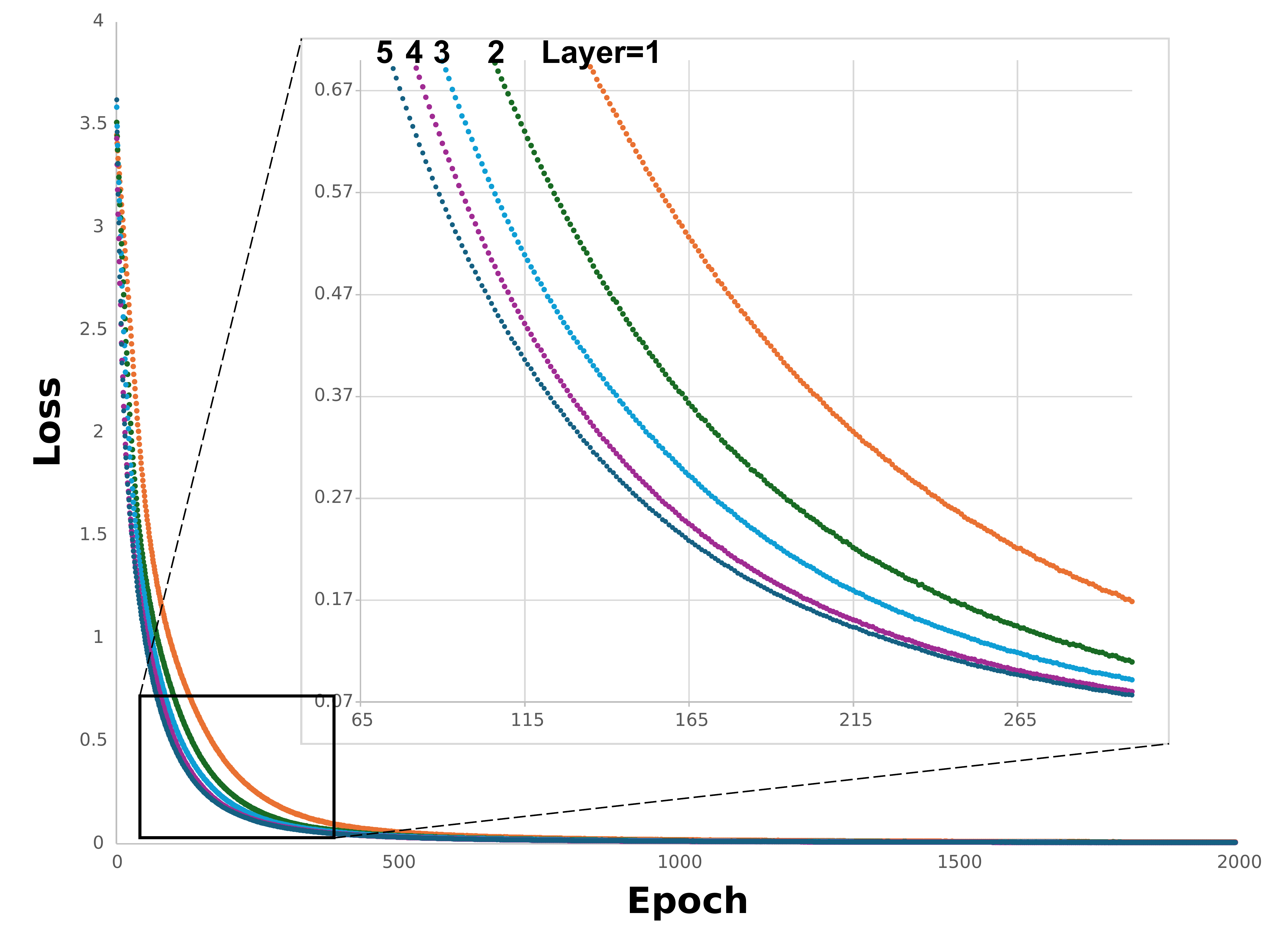}
    \caption{Increasing the number of embedding generator layers results in faster convergence, though the improvement is marginal.}
    \label{fig:layer}
    \end{subfigure}
    \hfill
    \begin{subfigure}[b]{0.49\textwidth}
        \centering
         \includegraphics[width=\linewidth]{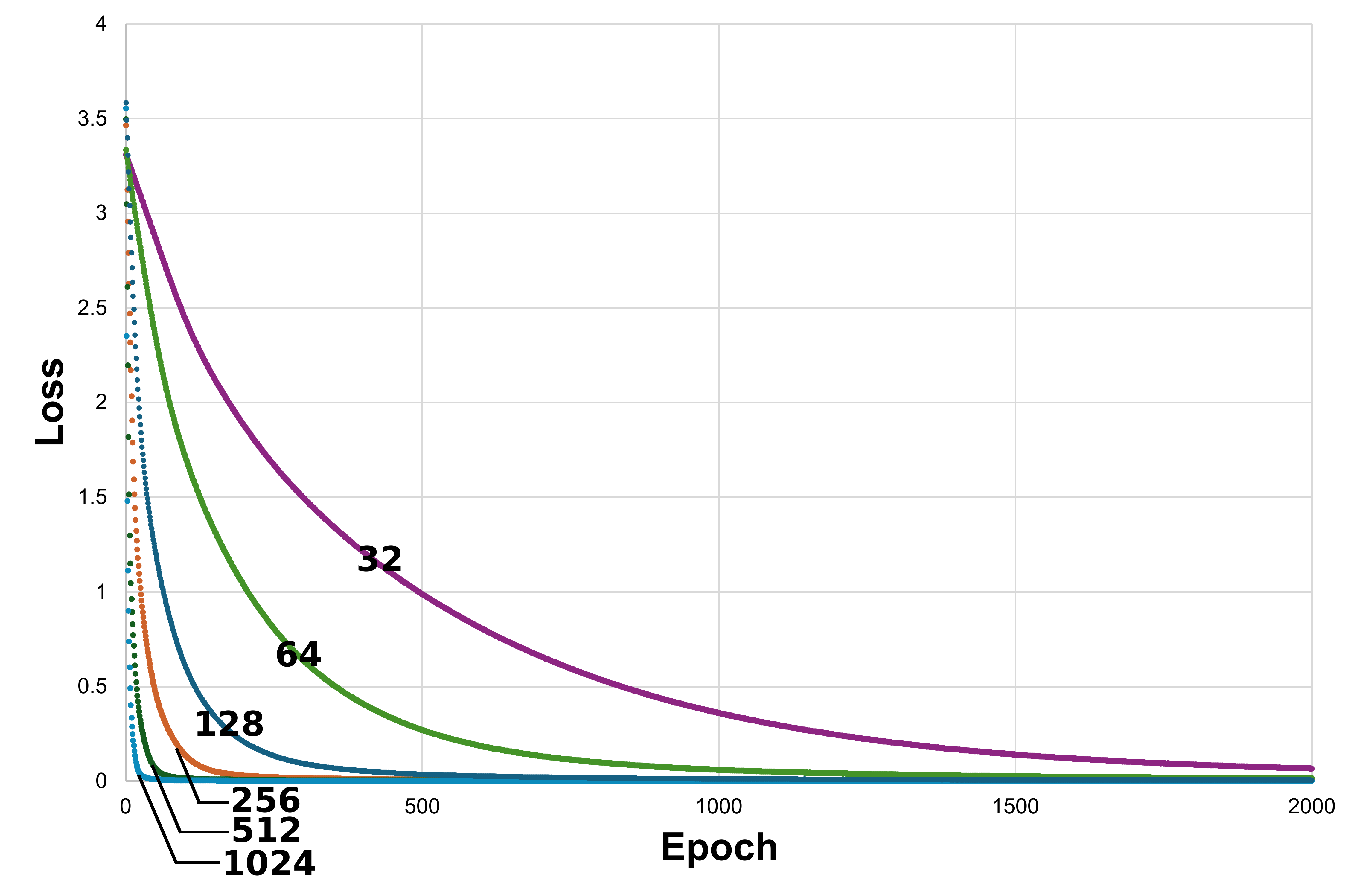}
        \caption{Larger embedding sizes also enhance convergence speed; however, smaller sizes (e.g., 32) may lead to invalid sequences.}
        \label{fig:embsize}
    \end{subfigure}
    \caption{Training trends in GPT-FT with varying embedding generator layer number and embedding size.}
    \label{fig:twoloss}
\end{figure}

\noindent\textbf{Parameter Sensitivity: number of embedding generator's layer.}
To validate the sensitivity of the embedding generator's layer count (see Section~\ref{sec:gpt}), we varied the number of layers from 1 to 5 and observed the training process and final performance. As shown in Figure~\ref{fig:layer}, the differences are minimal, with a trend of faster convergence as the number of layers increases. Based on the observation, we select a single layer to minimize inference time and model size.

\noindent\textbf{Parameter Sensitivity: GPT's embedding size}
To validate the sensitivity of GPT's embedding size, we varied it from 32 to 1024 and observed the training process and final performance. Figure~\ref{fig:embsize} shows that larger embedding sizes lead to faster convergence, but performance remains consistent for sizes between 64 and 1024. At 32, occasional invalid records are generated. Considering performance stability and model size, we select an embedding size of 64 for our experiments.

\section{Related Work}
Automated Feature Transformation (AFT) enhances feature spaces by applying mathematical operations to original features~\cite{chen2021techniques,kusiak2001feature}. Existing methods fall into three categories:  
1) \textbf{Expansion-reduction approaches}~\cite{kanter2015deep,khurana2016cognito,horn2019autofeat,lam2017one,khurana2016automating}, which expand the feature space via explicit~\cite{katz2016explorekit} or greedy~\cite{dor2012strengthening} transformations, then reduce it by selecting useful features. However, these approaches struggle with evaluating complex transformations, leading to subpar performance.  
2) \textbf{Evolution-evaluation approaches}~\cite{wang2022group,khurana2018feature,tran2016genetic,xiao2023traceable,zhu2022evolutionary,xiao2023traceableSIAM}, which integrate feature generation and selection in a closed-loop system optimized by evolutionary algorithms or reinforcement learning. While effective, they remain time-consuming and unstable due to reliance on discrete decision-making.  
3) \textbf{AutoML-based approaches}~\cite{chen2019neural,zhu2022difer}, inspired by AutoML’s success~\cite{elsken2019neural,li2021automl,he2021automl,karmaker2021automl}, formulate AFT as an AutoML task. However, these methods are limited by: 1) inability to produce high-order transformations; 2) unstable performance; and 3) reliance on discrete optimization.
MOAT~\cite{wang2023reinforcement} was introduced to address these deficiencies by framing AFT as a continuous optimization problem. However, MOAT utilized an LSTM model, which is considerably larger and less efficient compared to GPT. The experimental section demonstrates that GPT-FT outperforms MOAT, exhibiting a smaller parameter size and reduced inference time.

\section{Conclusion}
In this paper, we introduced GPT-FT, a novel framework for efficient automated feature transformation leveraging the capabilities of Generative Pre-trained Transformers (GPT) \cite{radford2018gpt}. By unifying transformation sequence reconstruction and model performance estimation within a single architecture, GPT-FT achieves a significant reduction in computational overhead and parameter size compared to existing methods. Through its four-stage process—transformation records collection, embedding space construction, gradient-ascent search, and autoregressive reconstruction, GPT-FT effectively addresses the scalability and efficiency challenges inherent in automated feature transformation.

Extensive experiments on benchmark datasets demonstrate that GPT-FT consistently outperforms state-of-the-art methods, achieving superior predictive performance while reducing inference time and model size. The robustness of GPT-FT across various machine learning models highlights its adaptability and practical utility for diverse applications. Furthermore, the integration of gradient-ascent search into the embedding space exemplifies the potential of continuous optimization techniques for feature engineering tasks.

Future work will extend GPT-FT to larger datasets and more complex feature spaces, while exploring advanced transformer architectures to enhance scalability. We also aim to integrate GPT-FT with privacy-preserving machine learning, where efficient encrypted computation could enable secure 
feature transformation \cite{gao2024secure} in sensitive
domains. Finally, adopting the evaluation benchmark \cite{liu2024stylerec} for sequence reconstruction and cross-domain prompt recovery will further strengthen robustness, underscoring GPT-FT’s potential to advance automated machine learning pipelines

\bibliographystyle{splncs04}
\bibliography{ref.bib}

\appendix
\section{Experiment}

\subsection{Experiment Platform Information}
All experiments were conducted on the Ubuntu 20.04.6 LTS operating system, Intel(R) Xeon(R) Silver 4114 CPU, and 4 NVIDIA TITAN RTX GPUs, with the framework of Python 3.8.5 and PyTorch 1.8.1.
\label{apx:pltfm}

\subsection{Hyperparameter Settings}
A single-layer embedding generator and a single-layer feed-forward network were employed for the text predictor and task classifier. The embedding size for all three models is 64. We utilized a single head for the self-attention block. In the training of GPT-FT, we established a batch size of 16, a learning rate of $1.31\times 10^{-5}$, and a trade-off hyperparameter $\alpha$ set at 0.133. To infer new transformation sequences, we utilized the top 42 records as the foundational seeds.
\vspace{-0.2cm}
\subsection{Experiment Details}
\label{apx:expdtl}
\vspace{-1.2cm}
\begin{table}[!ht]
    \centering
    \caption{Comparison of model parameter sizes between MOAT and GPT-FT across various datasets. The unit is Megabyte(MB).}
    \begin{tabular}{ccccc}
    \hline
        \textbf{Dataset} & \textbf{Samples} & \textbf{Features} & \textbf{MOAT} & \textbf{GPT-FT} \\ \hline
        Contraceptive Method Choice & 1473 & 9 & 0.42 & 0.21 \\ \hline
        Heart Disease & 303 & 13 & 0.20 & 0.10 \\ \hline
        Ozone Level Detection & 2536 & 72 & 0.64 & 0.31 \\ \hline
        Seeds & 210 & 7 & 0.32 & 0.16 \\ \hline
        Titanic & 891 & 11 & 0.31 & 0.15 \\ \hline
        Lymphography & 148 & 18 & 0.17 & 0.08 \\ \hline
        Amazon Employee & 32769 & 9 & 6.46 & 3.21 \\ \hline
        Wine Quality Red & 999 & 12 & 0.54 & 0.27 \\ \hline
        Wine Quality White & 4900 & 12 & 1.27 & 0.63 \\ \hline
        Tecator & 240 & 125 & 4.81 & 2.39 \\ \hline
        GeographicalOriginalofMusic & 1059 & 118 & 14.14 & 7.08 \\ \hline
        Jasmine & 2984 & 145 & 0.98 & 0.49 \\ \hline
        Libras move & 360 & 91 & 0.38 & 0.19 \\ \hline
        Bodyfat & 252 & 15 & 0.23 & 0.11 \\ \hline
        Weather & 366 & 12 & 0.92 & 0.45 \\ \hline
    \end{tabular}
    \label{tab:expsize}
\end{table}
\vspace{-1.7cm}
\begin{table}[!ht]
    \centering
    \caption{Comparison of model inference time between MOAT and GPT-FT across various datasets. The unit is second.}
    \begin{tabular}{ccccc}
    \hline
        \textbf{Dataset} & \textbf{Samples} & \textbf{Features} & \textbf{MOAT} & \textbf{GPT-FT} \\ \hline
        Contraceptive Method Choice & 1473 & 9 & 23.83 & 23.30 \\ \hline
        Heart Disease & 303 & 13 & 24.58 & 22.61 \\ \hline
        Ozone Level Detection & 2536 & 72 & 34.11 & 27.93 \\ \hline
        Seeds & 210 & 7 & 36.89 & 29.08 \\ \hline
        Titanic & 891 & 11 & 25.59 & 23.56 \\ \hline
        Lymphography & 148 & 18 & 27.21 & 24.44 \\ \hline
        Amazon Employee & 32769 & 9 & 32.19 & 23.43 \\ \hline
        Wine Quality Red & 999 & 12 & 27.35 & 23.94 \\ \hline
        Wine Quality White & 4900 & 12 & 25.22 & 23.54 \\ \hline
        Tecator & 240 & 125 & 67.22 & 39.42 \\ \hline
        GeographicalOriginalofMusic & 1059 & 118 & 71.10 & 41.37 \\ \hline
        Jasmine & 2984 & 145 & 57.57 & 43.02 \\ \hline
        Libras move & 360 & 91 & 31.83 & 29.14 \\ \hline
        Bodyfat & 252 & 15 & 34.67 & 28.35 \\ \hline
        Weather & 366 & 12 & 24.12 & 22.79 \\ \hline
    \end{tabular}
    \label{tab:expinfertime}
\end{table}

\end{document}